\documentclass[runningheads]{llncs}

\usepackage{graphicx} 
\usepackage{float} 
\usepackage{subfigure}
\usepackage{multirow}
\usepackage{amsmath}
\usepackage{amssymb}
\usepackage{cite}
\usepackage[marginal]{footmisc}

\usepackage[table,xcdraw]{xcolor}
\usepackage{mathptmx}
\begin{document}
\title{Improved Heatmap-based Landmark Detection}
%
\author{Huifeng Yao\inst{1}\and
Ziyu Guo\inst{1} \and
Yatao Zhang\inst{1} \and
Xiaomeng Li\inst{2}
}

\authorrunning{H. Yao et al.}
%
\institute{Shandong University, Shandong, China \and
Hong Kong University of Science and Technology, Hongkong, China
\email{\{yhfpro,gzypro\}@hotmail.com}}
\maketitle              
\footnote{Huifeng Yao and Ziyu Guo contributed equally to this work and should be considered joint first-authors.}

\begin{abstract}
Mitral valve repair is a very difficult operation, often requiring experienced surgeons. The doctor will insert a prosthetic ring to aid in the restoration of heart function. The location of the prosthesis' sutures is critical. Obtaining and studying them during the procedure is a valuable learning experience for new surgeons.
This paper proposes a landmark detection network for detecting sutures in endoscopic pictures, which solves the problem of a variable number of suture points in the images. Because there are two datasets, one from the simulated domain and the other from real intraoperative data, this work uses cycleGAN to interconvert the images from the two domains to obtain a larger dataset and a better score on real intraoperative data. This paper performed the tests using a simulated dataset of 2708 photos and a real dataset of 2376 images. The mean sensitivity on the simulated dataset is about 75.64 ± 4.48\% and the precision is about 73.62 ± 9.99\%. The mean sensitivity on the real dataset is about 50.23 ± 3.76\% and the precision is about 62.76 ± 4.93\%. The data is from the AdaptOR MICCAI Challenge 2021, which can be found at https://zenodo.org/record/4646979\#.YO1zLUxCQ2x.\par

\end{abstract}
\section{Introduction}
In mitral valve repair, the surgeon repairs part of the damaged mitral valve to allow the valve to fully close and stop leaking. The surgeon may tighten or reinforce the ring around a valve by implanting an artificial ring. The surgeon may place approximately 12 to 15 sutures on the mitral annulus~\cite{carpentier2011carpentier}. We need to know how sutures are placed because analyzing the pattern and distances between them can help us improve the quality of this surgery. Furthermore, the position of the sutures may aid the medico in learning how to perform this surgery by reconstructing it in a 3D virtual environment.\par
Deep learning methods have been widely used in the field of medical images. This task belongs to the landmark detection task in computer vision. In general, people mainly use the heatmap-based~\cite{payer2019integrating} method, coordinate regression method, and patch-based method. Payer et al.~\cite{payer2019integrating} used the SpatialConfiguration-Net which combines the local appearance of landmarks with their spatial configuration. Because the coordinate regression method is too difficult to converge and the patch-based method is difficult to distinguish adjacent points, we choose the heatmap-based method.\par
Many state-of-the-art heatmap-based deep learning methods focus on detecting fixed key points which are not suitable for our task. Stern et al.~\cite{stern2021heatmap} proposed a heatmap-based method to detecting a varying number of key points. Inspired by that, we present an improved heatmap-based method that can deal with a varying number of sutures and get better performance than that.\par
The data set is mainly split into two endoscopic sets. One is simulation data set and the other is real data set. Inspired by Engelhardt et al.~\cite{engelhardt2018improving}, we also implement the image to image translation to get more real data. We use the cycleGAN~\cite{zhu2017unpaired} network to do this task.\par
The work proposed a network to detect a varying number of landmarks and used the cycleGAN to translate images from two different domains. And we are participating within the scope of the AdaptOR challenge. \par

\section{Materials and methods}
\subsection{Data set}
Our data set comes from the AdaptOR challenge~\cite{sharan2021mutually}. The data set is mainly split into two endoscopic sets:\par

\noindent(1) Sim-Domain is the image acquired during simulating mitral valve repair on a surgical simulator. More information on the simulator can be found in ~\cite{engelhardt2019replicated} and ~\cite{engelhardt2019flexible}. The simulator dataset used for training consists of 2708 frames, which were extracted from 10 surgeries. We divide it into 5 fold. To prevent data leakage, dataset splitting was always carried out on the level of the surgeries. \par

\noindent(2) Intraop-Domain is the Intraoperative endoscopic data from real minimally invasive mitral valve repair. Since the intraoperative dataset consists of 2376 frames extracted from 4 simulated surgeries, we split it into 4 fold with each surgery comprising one fold. \par

The Label of this data set is stored in the format of a JSON file. 
In addition, the data splitting is shown in Table 1. \par

\begin{table}[H]
\small
\centering
\caption{Data set.}
\setlength{\tabcolsep}{4.5mm}
\begin{tabular}{|c|c|c|c|c|c|c|}
\hline
                     &            & \multicolumn{5}{c|}{Number of frames} \\ \hline
Domain               & Split      & $f$1    & $f2$    & $f3$    & $f4$    & $f5$    \\ \hline
\multirow{2}{*}{Sim} & Train      & 2246  & 2144  & 1960  & 2174  & 2308  \\ \cline{2-7} 
                     & Validation & 462   & 564   & 748   & 534   & 400   \\ \hline
\multirow{2}{*}{Intraop}  & Train      & 1582  & 1852  & 2004  & 1690  & -     \\ \cline{2-7} 
                     & Validation & 794   & 524   & 372   & 686   & -     \\ \hline
\end{tabular}
\end{table}

\subsection{Outline of the proposed method }
We have a lot of simulated data, but we don't have enough real data. So the first step is the image to image translation. We use a cycleGAN to convert simulated data to real data in order to obtain more real data, which will help our model score higher on the real dataset. The second step is to generate the heatmap. Unlike other tasks about landmark detection, which use one channel for each landmark, we do not have fixed points in this task. So we generate all the points in one channel. And each of them is a 2D Gaussian kernel. We do some augmentation for both the original image and heatmap. Then the enhanced images would be the input of the U-net-based~\cite{unet} network. The corresponding heatmap would be the label of the image. Then, we use the Otsu~\cite{otsu1979threshold} to get the thresholding image. We also Use the open operation to remove the noise in the image and make the binarized area smoother. Finally, we use the cutting method to separate very close points and the centroid of each region is taken as the final result. All of these are shown in Fig.1.\par

\begin{figure}[H] 
\centering 
\includegraphics[width=\textwidth]{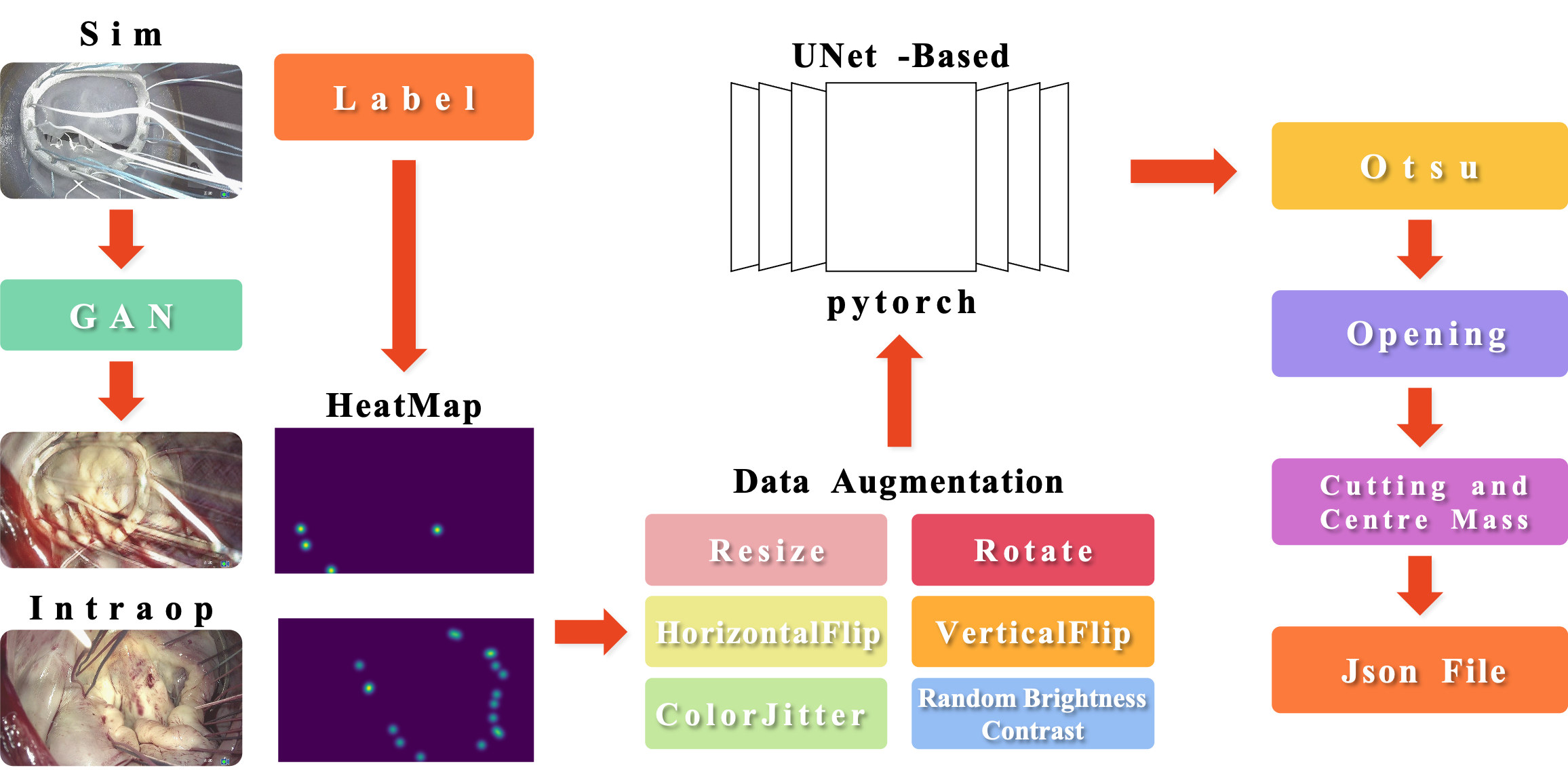}
\caption{Outline} 
\label{Fig.main1} 
\end{figure}

\subsection{Pre-processing} 
\subsubsection{image to image GAN}
In this task, our datasets come from two domains, one is the simulation domain and the other is the Intraop domain. The data set of the Intraop domain is smaller than the data set of the simulation domain. We decided to transform the simulation domain data into Intraop domain data to get a higher score on the Intraop domain. We introduced cycleGAN to solve this problem.\par
The cycleGAN has two mapping functions, as shown in Fig.2, one is G and the other is F. G transforms the image of X domain into the image of Y domain, and F transforms the image of Y domain into the image of X domain. Two discriminators identify the real domain image and the generated image.

\begin{figure}[H] 
\centering 
\includegraphics[width=0.6\textwidth]{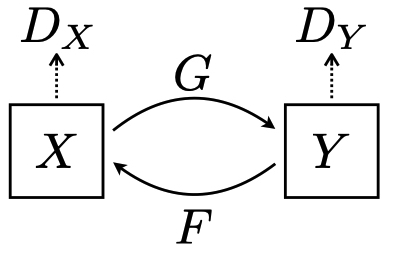}
\caption{cycleGAN} 
\label{Fig.main2} 
\end{figure}

Applied to this task, the overall flow is shown in Fig.3. This diagram only shows the process from the simulation domain to the Intraop domain and vice versa, which is not shown here.\par

\begin{figure}[H] 
\centering 
\includegraphics[width=1\textwidth]{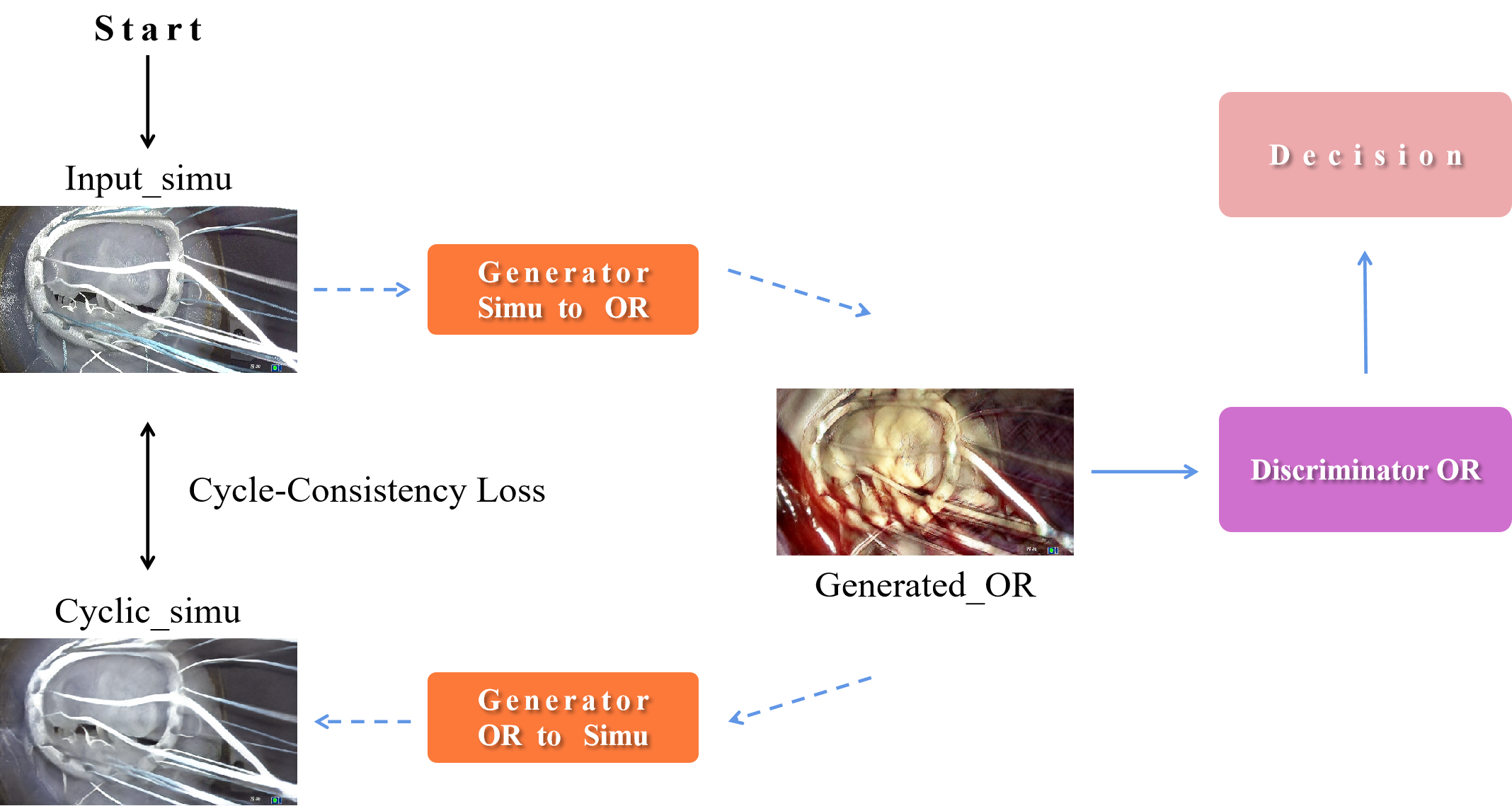}
\caption{cycleGAN in this task} 
\label{Fig.main3} 
\end{figure}

\subsubsection{heatmap}
Unlike the traditional landmark detection method, we do not generate a heatmap for each point but generate all the points onto the same heatmap. Because in other tasks, the number of feature points is fixed, while in our task, the number of feature points varies with the image, ranging roughly from 0 to 15.\par
Each of our points is a 2D Gaussian kernel, and a variable number of points make up this heatmap, which will be used as the model's label. The heatmap is shown in Fig.4.

\begin{figure}[H]
\centering 
\includegraphics[width=1\textwidth]{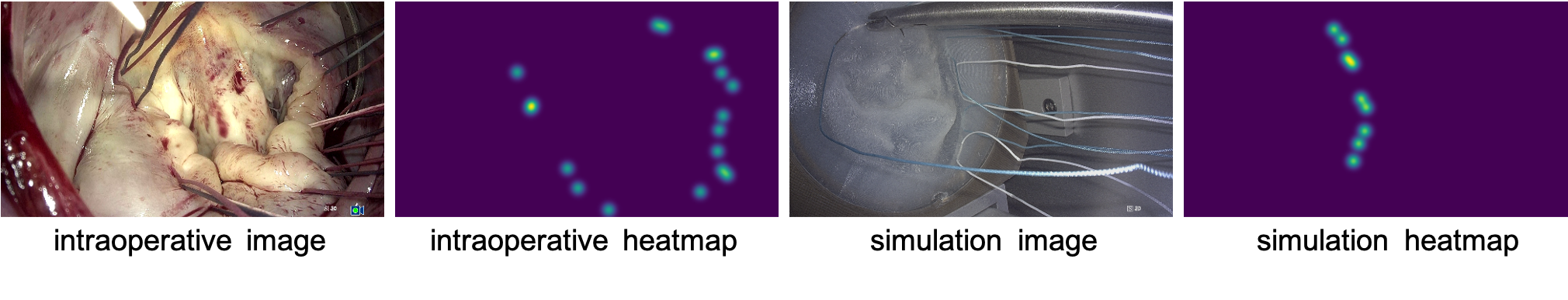}
\caption{heatmap} 
\label{Fig.main4} 
\end{figure}

\subsubsection{data augmentation}
During training the images are randomly augmented using Albumentations functions: horizontally and vertically with a probability of 50\%, rotation of  $\pm$ 40°, ColorJitter with a probability of 50\%, RandomBrightnessContrast with a probability of 50\%.

\subsection{Point detection}
This work uses a U-Net-based architecture with a depth of 5. After each 3×3-convolution, batch normalization is applied. The first convolutional layer has 16 filter maps, while the bottleneck layer has 512 filter maps. We choose the Resnext~\cite{xie2017aggregated} network as our encoder. We don't have an activation function after the final 1 × 1-convolutional layer while training, but we apply the sigmoid function when we predict the heatmap. The loss function is dice loss. \par 
The input images are RGB images with 3 channels. One channel output is the heatmap. The heatmap becomes the real output point after a series of subsequent operations.

\subsection{Post-processing}

\subsubsection{Otsu}
The maximum between-class variance method is a nonparametric and unsupervised method of automatic threshold selection for picture segmentation. According to the gray characteristics of the image, the image is divided into background and objects. Among them, the greater the variance between the background and objects shows that the difference between the two parts of the image is also greater. This method calculates the relationship between the average gray level between background pixels and foreground pixels and their proportion in the whole image, so as to obtain the global threshold when the image segmentation effect is the best, and finally segment the image according to this value.\par

\subsubsection{Opening}
We discovered that the network's predicted images were connected together in blocks that should have been separated after Otsu. The open operation is used to separate them. This also smoothes the edges of the segmented blocks and removes some of the noise.

\subsubsection{Centre mass and Cutting}After the opening process, we identify the centroid of each segmented block in the output image, and these are regarded the final predicted points, but we discovered that the form of some of these blocks compared to the circle generated by the point in the heatmap image is somewhat irregular. Therefore, we assess whether a segmentation block should be clipped depending on whether the area of each segmented block in an output image exceeds the average value of all its segmented blocks. Then, based on the height and width of the segmented blocks' bounding box, decide the cutting direction. The cutting point is the centroid of the segmented blocks that need to be sliced. Cutting is done in the x-axis direction if the bounding box's height is higher than its width. If the bounding box's height is less than its width, the cutting is done using Cut in the y-axis direction.
We recalculate the centroid of the partitioned block after cutting as the output points and save them in JSON files. \par
The example of post-processing is shown in Fig.5.\par

\begin{figure}[H]
\centering 
\includegraphics[width=1\textwidth]{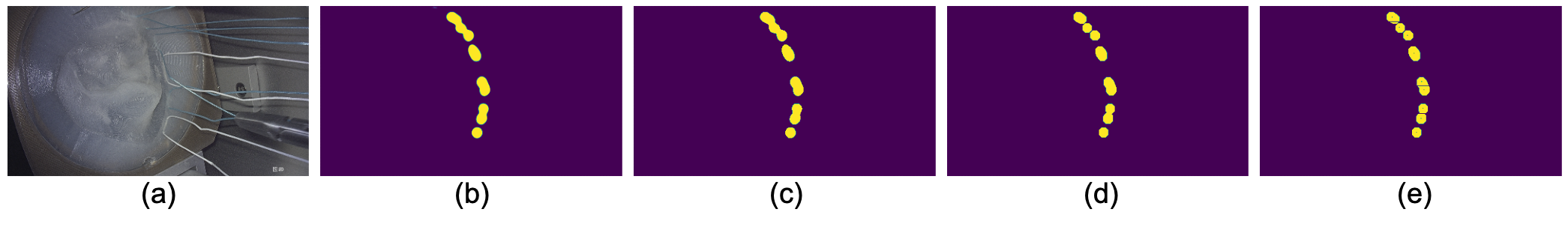}
\caption{Example of post-processing. (a) input, (b) predict, (c) Otsu, (d) opening, (e) Centre mass and Cutting}  
\label{Fig.main5} 
\end{figure}

\subsection{Evaluation}
A point detection is considered successful if the centres of mass of ground truth and prediction are less than 6 pixels apart. On an image of size 512 × 288, this radius roughly corresponds to the thickness of a suture when it enters the tissue. Every matched point from the produced mask is considered a true positive (TP). Predicted points that could not be matched to any ground truth point are defined as false positives (FP) and all ground truth points without a corresponding point in the produced mask are false negatives (FN). Precision and sensitivity are computed over all landmarks. And F1-score presents the harmonic mean of precision and sensitivity.

\begin{gather}
    Precision = \frac{TP}{TP + FP }\\
	Sensitivity = \frac{TP}{TP + FN}\\
	F1 = \frac{2*Precision*Sensitivity}{Precision+Sensitivity}
\end{gather}

\section{Results}
There are some visual examples in Fig.6(a) and Fig.6(b). 
\begin{figure}[H]
\centering 
\includegraphics[width=1\textwidth]{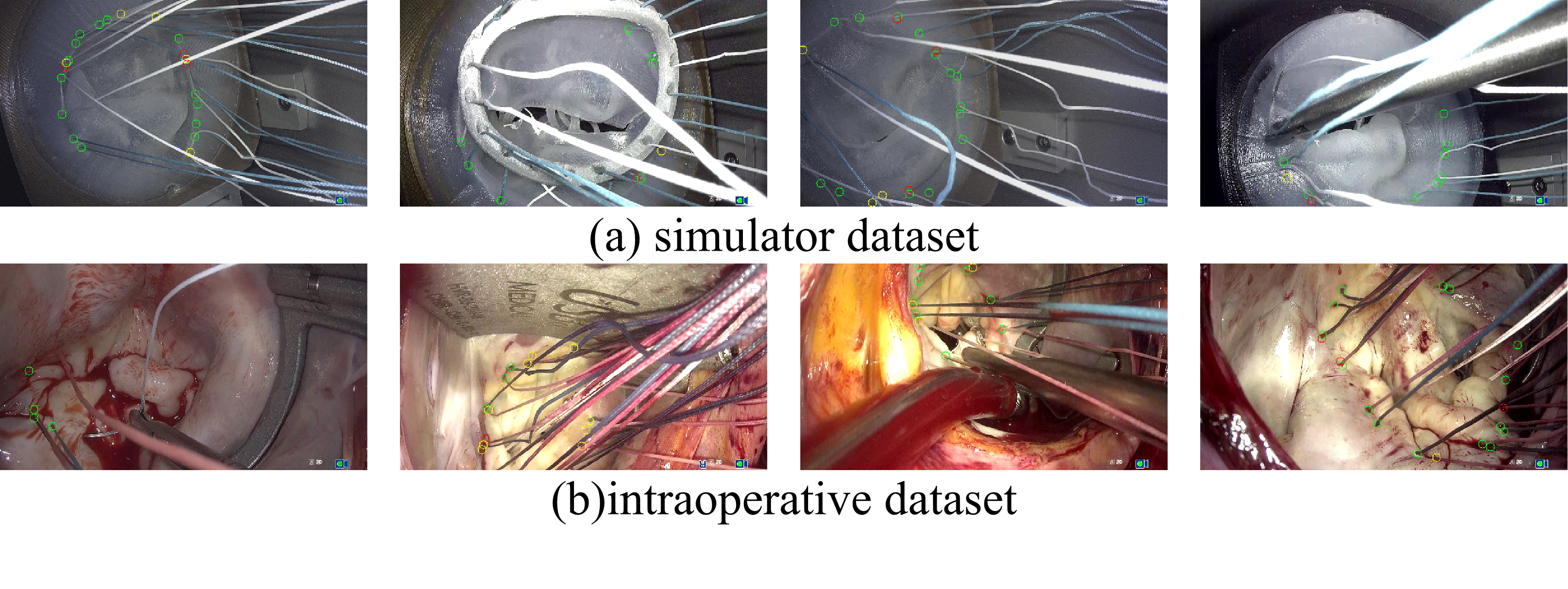}
\caption{Example results of two domain. The green circles are true positives(TP). The red circles show false positives(FP). The yellow circles represent false negatives (FN).} 
\label{Fig.main6} 
\end{figure}

The result of the simulation domain is shown in Table 2, and the result of the intraop domain is shown in Table 3.

The baseline results come from this paper~\cite{stern2021heatmap}. We can not calculate the standard deviation of the baseline F1 score since the baseline does not give experimental data for F1 score.\par

\begin{table}[H]
\centering
\caption{Simu result.}\label{tab2}
\setlength{\tabcolsep}{2.5mm}
\begin{tabular}{cccccccc} 
\hline
\multicolumn{8}{c}{Cross-validation result on Simu data}                                                  \\ 
\hline
Metric                       & Model    & $f1$    & $f2$    & $f3$    & $f4$    & $f5$    & $\mu$ ± $\sigma$         \\ 
\hline
\multirow{2}{*}{Precision}   & Baseline & -     & -     & -     & -     & -     & 81.50 ± 5.77             \\
                             & Ours     & 84.37 & 54.79 & 76.84 & 74.18 & 77.89 & 73.62 ± 9.99             \\ 
\hline
\multirow{2}{*}{Sensitivity} & Baseline & -     & -     & -     & -     & -     & 61.60 ± 6.11             \\
                             & Ours     & 79.63 & 72.25 & 68.64 & 80.20 & 77.48 & \textbf{ 75.64 ± 4.48 }  \\ 
\hline
\multirow{2}{*}{F1 score}    & Baseline & -     & -     & -     & -     & -     & 69.78                    \\
                             & Ours     & 81.94 & 62.33 & 72.51 & 77.07 & 77.69 & \textbf{ 74.31 ± 6.69 }  \\
\hline
\end{tabular}
\end{table}

\begin{table}[H]
\centering
\caption{Intra result.}\label{tab3}
\setlength{\tabcolsep}{3.4mm}
\begin{tabular}{ccccccc} 
\hline
\multicolumn{7}{c}{Cross-validation result on Intra data}                                        \\ 
\hline
Metric                       & Model    & $f1$    & $f2$    & $f3$    & $f4$    & $\mu$ ± $\sigma$         \\ 
\hline
\multirow{2}{*}{Precision}   & Baseline & -     & -     & -     & -     & 66.68 ± 4.67             \\
                             & Ours     & 62.24 & 67.35 & 54.92 & 66.54 & 62.76 ± 4.93             \\ 
\hline
\multirow{2}{*}{Sensitivity} & Baseline & -     & -     & -     & -     & 24.45 ± 5.06             \\
                             & Ours     & 51.81 & 54.44 & 44.22 & 50.45 & \textbf{ 50.23 ± 3.76 }  \\ 
\hline
\multirow{2}{*}{F1 score}    & Baseline & -     & -     & -     & -     & 35.78                    \\
                             & Ours     & 56.56 & 60.22 & 48.99 & 57.38 & \textbf{ 55.79 ± 4.15 }  \\
\hline
\end{tabular}
\end{table}

\noindent As shown in Tables 2 and 3, while our precision is lower than the baseline, our sensitivity is much higher. As a result, when comparing F1 scores, our method outperforms the baseline on both the simulation and intraop domains. Because the images in the intraop domain have more interference factors and less data, the recognition effect of the two methods in the intraop domain is slightly inferior to that of the simulation domain.

\section{Conclusions}
We present a novel method for predicting multiple key points in endoscopic images in this paper. Our method differs from traditional key point detection methods, which have a fixed number of prediction key points. Our method can detect multiple key points at the same time, significantly reducing detection time and model calculation. We also introduce cycleGAN, which can interconvert images from two domains to create a larger dataset. Our results outperform the baseline as well as other related methods after many repeated and rigorous experiments.

%
%
%

\bibliographystyle{splncs04}
\bibliography{my}
%
    \nocite{carpentier2011carpentier}
    \nocite{payer2019integrating}
    \nocite{stern2021heatmap}
    \nocite{engelhardt2018improving}
    \nocite{engelhardt2019replicated}
    \nocite{engelhardt2019flexible}
    \nocite{zhu2017unpaired}
    \nocite{unet}
    \nocite{otsu1979threshold}
    \nocite{xie2017aggregated}
    \nocite{sharan2021mutually}

\end{document}